\documentclass[aps,pre,reprint,amsmath,amssymb,floatfix]{revtex4-2}
\usepackage{bm}
\usepackage{graphicx}
\usepackage{xcolor}
\usepackage[hidelinks]{hyperref}
\begin{document}
\title{Learning Koopman Operators for Coupled Systems via Information on Governing Equations of Subsystems}
\author{Tatsuya Naoi}
\email{t.naoi.982@ms.saitama-u.ac.jp}
\author{Jun Ohkubo}
\email{johkubo@mail.saitama-u.ac.jp}
\affiliation{Graduate School of Science and Engineering, Saitama University, Sakura, Saitama, 338-8570, Japan}
\begin{abstract}
Nonlinear coupled systems are ubiquitous in science and engineering. The analysis and modeling of such systems are challenging due to their high dimensionality and complex interactions among subsystems. In recent years, operator-theoretic methods based on the Koopman operator have attracted attention as a powerful tool for analyzing and modeling nonlinear dynamical systems. Extended dynamic mode decomposition (EDMD) is one of the most popular methods for approximating the Koopman operator. However, EDMD is a purely data-driven method, and it may be unstable and inaccurate for coupled systems under limited data availability. In this paper, we propose a method to construct a finite-dimensional Koopman matrix for coupled systems using the differential equations governing each subsystem. By regularizing the Koopman matrix toward a prior constructed from the subsystem equations, the proposed method improves prediction accuracy when only a limited number of snapshot pairs are available. We also demonstrate its effectiveness through numerical experiments on coupled oscillator systems.
\end{abstract}
\maketitle

\section{Introduction}
\label{sec:introduction}
Coupled nonlinear dynamical systems are prevalent in various fields, including physics, engineering, biology, and social sciences. Examples of coupled systems include coupled oscillators, power grids, and multi-agent networks. Analyzing and modeling such systems is challenging due to their high dimensionality and complex interactions among subsystems.

Coupled systems have been extensively studied in physics. One example is synchronization.\cite{strogatz2000kuramoto,fink2000three,acebron2005kuramoto,BOCCALETTI2006175,ARENAS200893,mirzaei2023synchronization} While synchronization provides an important perspective on the collective behavior of coupled systems, it does not by itself address how to construct compact and predictive representations of coupled nonlinear dynamical systems. As a result, a variety of data-driven methods have been developed to learn predictive models of coupled nonlinear dynamical systems.\cite{Stephan2015data,xiao2021predicting,ghadami2022data} System identification methods have been proposed to infer the governing equations from data.\cite{Brunton2016sindy,rudy2017data}

Among these data-driven methods, operator-theoretic methods based on the Koopman operator have attracted increasing attention.\cite{koopman1931hamiltonian} They provide a linear representation of nonlinear dynamics in an infinite-dimensional function space.\cite{brunton2022modern} For example, extended dynamic mode decomposition (EDMD) is a data-driven method for approximating the Koopman operator.\cite{williams2015data} Although the Koopman operator is generally infinite-dimensional, its spectral properties enable modal decomposition and stability analysis of nonlinear dynamical systems in terms of eigenvalues, eigenfunctions, and modes.\cite{mezic2005spectral,mezic2013spectral,susuki2014nonlinear,mauroy2016global} The linear representation of nonlinear dynamics also enables the application of various linear control techniques.\cite{proctor2016dynamic,brunton2016control}

The application of Koopman operator theory to coupled systems has also been studied.\cite{schlosser2022sparsity,tellez2022data,guo2025modularized} These studies have proposed more effective approaches to construct finite-dimensional Koopman approximations for coupled systems by leveraging their structure and interactions. In addition, Koopman-based analysis has led to investigations of the phase-space structure of coupled oscillator systems and the synchronization of oscillators.\cite{wang2021probing,hu2020koopman} However, EDMD-based approaches can face significant challenges when applied to nonlinear systems with multiple time scales and high dimensionality.\cite{dylewsky2019dynamic} Moreover, learning a high-dimensional Koopman approximation generally requires a large number of snapshot pairs. This motivates the development of data-efficient methods that exploit available prior information.

The incorporation of prior information and physical constraints into the
learning process can improve the accuracy and physical consistency of
Koopman operator approximations. For example, structural information,
such as interaction topology, has been embedded into Koopman-based
learning methods.\cite{tellez2022data,guo2025modularized} Other studies
have incorporated conservation laws and related physical properties into
EDMD-based approximations.\cite{colbrook2023mpedmd,baddoo2023physics,chen2025online}

These studies demonstrate the benefit of incorporating structural or
physical information, but they primarily consider interaction topology,
invariance properties, or physical constraints. A distinct source of prior
information is the governing dynamics of individual subsystems. In many
applications, the governing equation of the complete coupled system,
especially the coupling terms, may be unavailable, whereas the differential
equations governing the individual subsystems are known. Such subsystem
models can provide a physically meaningful prior for learning the Koopman
operator of the full coupled system. By incorporating these subsystem models as prior information, it may be possible to reduce the amount of snapshot data required to construct a finite-dimensional Koopman approximation for the full coupled system.

Several studies have demonstrated data-efficient learning of
Koopman operator approximations by incorporating governing equations as prior
information, thereby reducing the reliance on snapshot data.\cite{ristich2025physics,ohta2025integrated}
However, to the best of our knowledge, the use of known subsystem
equations as prior information for learning a global Koopman approximation for
an otherwise unknown coupled system has not been systematically studied.

In this paper, we propose a data-efficient method to construct a finite-dimensional Koopman approximation for coupled systems by utilizing the differential equations governing each subsystem. Rather than learning the entire Koopman matrix solely from data, the proposed method constructs a physics-informed prior from the subsystem equations and learns only the coupling-induced correction from snapshot data. The proposed method consists of four main steps. First, we derive Koopman matrices for the individual subsystems from their governing differential equations using the methods proposed in Refs.~\cite{ohkubo2019duality,ohkubo2022numerical}. We then construct a tensor-product prior for the global Koopman matrix from these subsystem-level approximations. Third, the prior is projected onto the truncated global dictionary. Finally, the prior is incorporated as the center of a regularized EDMD problem, through which the coupling-induced correction is learned from snapshot data of the full coupled system.

The rest of this paper is organized as follows. In Sect.~\ref{sec:preliminaries}, we review coupled systems, Koopman operator theory, and EDMD. In Sect.~\ref{sec:proposed_method}, we describe the proposed method to construct a finite-dimensional Koopman approximation for coupled systems using the differential equations governing each subsystem. In Sect.~\ref{sec:numerical_experiments}, we present numerical experiments on coupled oscillator systems to demonstrate the effectiveness of the proposed method. Finally, in Sect.~\ref{sec:conclusion}, we conclude the paper and discuss future research directions.

\section{Preliminaries}
\label{sec:preliminaries}
\subsection{Coupled systems}
\label{sec:coupled_systems}

We consider a continuous-time coupled dynamical system composed of $N$ subsystems:
\begin{widetext}
\begin{align}
    \dot{\bm{z}}_{i}(t)=
    \bm{f}_{i}\left(\bm{z}_{i}(t)\right)
    +
    \sum_{j\in\mathcal{N}_{i}}
    c_{ij}
    \bm{g}_{ij}\left(
        \bm{z}_{i}(t),
        \bm{z}_{j}(t)
    \right),
    \quad i=1,\ldots,N.
    \label{eq:coupled_system}
\end{align}
\end{widetext}
Here, $\bm{z}_{i}(t)\in\mathbb{R}^{D_{i}}$ denotes the state of the $i$-th subsystem, and $D_{i}$ is the dimension of the $i$-th subsystem. The state vector of the entire coupled system is defined by
\begin{equation}
    \bm{z}(t) =
    \begin{bmatrix}
        \bm{z}_{1}(t)^\top &
        \bm{z}_{2}(t)^\top &
        \ldots &
        \bm{z}_{N}(t)^\top
    \end{bmatrix}^{\top}
    \in\mathbb{R}^{D},
    \label{eq:global_state_vector}
\end{equation}
where $D=\sum_{i=1}^{N}D_{i}$ is the total dimension of the coupled system.
The function $\bm{f}_{i}:\mathbb{R}^{D_i}\to\mathbb{R}^{D_i}$ describes the intrinsic dynamics of the $i$-th subsystem, whereas $\bm{g}_{ij}:\mathbb{R}^{D_i}\times\mathbb{R}^{D_j}\to\mathbb{R}^{D_i}$ represents the influence of the $j$-th subsystem on the $i$-th subsystem. The coefficient $c_{ij}$ denotes the coupling strength, and $\mathcal{N}_{i}$ is the set of subsystems coupled to the $i$-th subsystem.

Several Koopman-based methods for coupled systems explicitly exploit interconnection structures, such as the interaction topology $\mathcal{N}_{i}$, to construct localized or modular predictive models.\cite{tellez2022data,guo2025modularized} In contrast, this study considers a complementary setting in which the complete governing equations of the coupled system are unavailable. In particular, the coupling functions $\bm{g}_{ij}$, the coupling strengths $c_{ij}$, and possibly the
interaction topology are not assumed to be known.

We instead assume that the intrinsic subsystem dynamics $\bm{f}_{i}$ are known a priori. This situation arises when reliable mathematical models are available for the individual components, whereas their interconnection mechanisms are unknown or difficult to model. The objective of this study is to use the known subsystem dynamics as prior information for learning a Koopman approximation of the full coupled system from snapshot data.

\subsection{Koopman operator theory}
\label{sec:koopman_operator_theory}
The Koopman operator is a linear operator that describes the evolution of observables in dynamical systems.
We consider the following discrete-time dynamical system:
\begin{align}\label{eq:dynamical system}
    \bm{x}_{m+1} = \bm{F}(\bm{x}_m),\quad m=0,1,2,\ldots,
\end{align} 
where $\bm{x}_{m}\in\mathbb{R}^{D}$ denotes the state vector at discrete time $m$, and $\bm{F}:\mathcal{M}\to\mathcal{M}$ is the nonlinear state-transition map.
In this subsection, we use the generic notation $\bm{x}_m$ to formulate Koopman operator theory. When applying the formulation to the coupled system in Eq.~\eqref{eq:coupled_system}, $\bm{x}_m$ corresponds to the full state vector $\bm{z}(m\Delta t)$ sampled at discrete time intervals
$\Delta t$. Here, we consider a deterministic system for simplicity. 

We also consider an observable function $\phi:\mathcal{M}\to\mathbb{C}$. The Koopman operator $\mathcal{K}$ acts on the observable function $\phi$ as follows:
\begin{align}\label{eq:koopman}
    \mathcal{K}\phi = \phi\circ\bm{F},
\end{align} 
where $\circ$ denotes the composition of $\phi$ with $\bm{F}$. 
Equations~\eqref{eq:dynamical system} and~\eqref{eq:koopman} lead to
\begin{align}\label{eq:koopman2}
    \phi(\bm{x}_{m+1}) = (\mathcal{K}\phi)(\bm{x}_m).
\end{align}
It is easy to confirm the linearity of the Koopman operator: for any constants $c_1,c_2\in\mathbb{C}$ and observable functions $\phi_1,\phi_2$, we have
\begin{align}\label{eq:linearity}
    \left\{\mathcal{K}(c_{1}\phi_{1}+c_{2}\phi_{2})\right\}(\bm{x}_{m}) &=(c_{1}\phi_{1}+c_{2}\phi_{2})(\bm{x}_{m+1}) \notag \\
    &= c_1\phi_1(\bm{x}_{m+1}) + c_2\phi_2(\bm{x}_{m+1}) \notag \\
    &= c_1(\mathcal{K}\phi_1)(\bm{x}_m) + c_2(\mathcal{K}\phi_2)(\bm{x}_m).
\end{align}

The Koopman operator $\mathcal{K}$ is infinite-dimensional because it acts on elements in the function space. Therefore, it is necessary to approximate the Koopman operator $\mathcal{K}$ by the Koopman matrix $K$ in a finite-dimensional subspace for practical applications. 

To describe the Koopman matrix, we introduce a dictionary, which is defined as follows:
\begin{align}\label{eq:dictionary}
    \bm{\Psi}(\bm{x}) = [\psi_1(\bm{x}),\psi_2(\bm{x}),\ldots,\psi_{N_\mathrm{dic}}(\bm{x})]^{\top},
\end{align}
where $\psi_{k}:\mathcal{M}\to\mathbb{C}$ is the $k$-th dictionary function, and $N_\mathrm{dic}$ is the number of dictionary functions. 
Then, the observable function $\phi$ is represented as a linear combination of the dictionary functions as follows:
\begin{align}\label{eq:observable}
    \phi(\bm{x}) = \sum_{k=1}^{N_\mathrm{dic}}c_{k}\psi_{k}(\bm{x})=\bm{c}^{\top}\bm{\Psi}(\bm{x}),
\end{align}
where $\bm{c}\in\mathbb{C}^{N_\mathrm{dic}}$ is a coefficient vector. Combining Eqs.~\eqref{eq:koopman2} and~\eqref{eq:observable}, we have
\begin{align}\label{eq:koopman3}
    (\mathcal{K}\phi)(\bm{x}_m) &= \bm{c}^{\top}(\mathcal{K}\bm{\Psi})(\bm{x}_m). 
\end{align}
Note that $\bm{c}$ is time-independent. Hence, instead of considering the action of the Koopman operator on the observable function $\phi$, it is sufficient to consider the action on the dictionary as follows:
\begin{align}\label{eq:koopman on dictionary}
    \bm{\Psi}(\bm{x}_{m+1}) = \mathcal{K}\bm{\Psi}(\bm{x}_m) \simeq K\bm{\Psi}(\bm{x}_m),
\end{align}
which leads to the Koopman matrix $K$.

\subsection{Brief summary of EDMD}
\label{sec:edmd}
In Ref.~\cite{williams2015data}, Williams \textit{et al.} proposed EDMD as a method for deriving the Koopman matrix $K$ from data.
Here, we consider a single time-series dataset $\{\bm{x}_{1},\bm{x}_{2},\ldots,\bm{x}_{M+1}\}$, although it is sufficient to use snapshot pairs rather than a single time series. The snapshot pairs $(\bm{x}_{m},\bm{x}_{m+1})$ correspond to $(\bm{x}_{m},\bm{y}_{m})$ in the following notation. The least-squares problem with the cost function
\begin{align}
    J = \sum_{m=1}^{M}\left\|\bm{\Psi}(\bm{y}_{m}) - K\bm{\Psi}(\bm{x}_{m})\right\|_{2}^{2},
    \label{eq:edmd_objective}
\end{align}
leads immediately to the Koopman matrix $K$.
Here, we define the following two matrices:
\begin{align}
    \bm{\Psi}_{X} &= \left[\bm{\Psi}(\bm{x}_{1}),\bm{\Psi}(\bm{x}_{2}),\ldots,\bm{\Psi}(\bm{x}_{M})\right], \\
    \bm{\Psi}_{Y} &= \left[\bm{\Psi}(\bm{y}_{1}),\bm{\Psi}(\bm{y}_{2}),\ldots,\bm{\Psi}(\bm{y}_{M})\right].
    \label{eq:edmd_snapshot_matrices}
\end{align}
Then, the Koopman matrix $K$ is given by
\begin{align}
    K = AG^{+},
    \label{eq:edmd_solution}
\end{align}
where
\begin{align}\label{eq:matrices Q P}
    A = \bm{\Psi}_{Y}\bm{\Psi}_{X}^{\top},\quad G = \bm{\Psi}_{X}\bm{\Psi}_{X}^{\top},
\end{align}
and $G^{+}$ is the Moore-Penrose pseudoinverse of $G$. 

\subsection{Derivation of Koopman matrix from differential equations}\label{sec:koopman_matrix_derivation_from_differential_equations}
Following Refs.~\cite{ohkubo2019duality,ohkubo2022numerical}, we describe a method for deriving a Koopman matrix directly from ordinary differential equations. The derivation is based on the duality between the Perron--Frobenius operator, which evolves probability densities, and the Koopman operator, which evolves observables. Since the full derivation is rather involved, we summarize only the essential steps needed in this study; for details, see Refs.~\cite{ohkubo2019duality,ohkubo2022numerical}.

We consider the following deterministic ordinary differential equation:
\begin{equation}
    \frac{d\bm{x}(t)}{dt}=\bm{f}(\bm{x}(t)),
    \qquad \bm{x}(0)=\bm{x}_0,
    \label{eq:continuous_dynamical_system}
\end{equation}
where $\bm{x}(t)\in\mathbb{R}^{D}$ and $\bm{f}:\mathbb{R}^{D}\to\mathbb{R}^{D}$ is a nonlinear vector field.

Let $p(\bm{x},t)$ denote a probability density transported by the flow of Eq.~\eqref{eq:continuous_dynamical_system}. Its time evolution is governed by the Liouville equation
\begin{equation}
    \frac{\partial p(\bm{x},t)}{\partial t}
    =\mathcal{L}^{*}p(\bm{x},t),
    \label{eq:liouville_equation}
\end{equation}
where the infinitesimal generator of the Perron--Frobenius operator is
\begin{equation}
    \left(\mathcal{L}^{*}p\right)(\bm{x})
    =-\sum_{i=1}^{D}\frac{\partial}{\partial x_i}
    \left[f_i(\bm{x})p(\bm{x})\right].
    \label{eq:liouville_generator}
\end{equation}

For the deterministic initial condition $\bm{x}_0$, the initial density is $\delta(\bm{x}-\bm{x}_0)$. Hence, for an observable $\phi$, its value along the trajectory can be written as
\begin{align}
    \phi(\bm{x}(t))
    &=\int_{\mathbb{R}^{D}}\phi(\bm{x})
    \left(e^{t\mathcal{L}^{*}}\delta(\bm{x}-\bm{x}_0)\right)(\bm{x})
    \,d\bm{x} \notag\\
    &=\int_{\mathbb{R}^{D}}\left(e^{t\mathcal{L}}\phi\right)(\bm{x})
    \delta(\bm{x}-\bm{x}_0)\,d\bm{x} \notag\\
    &=\left(e^{t\mathcal{L}}\phi\right)(\bm{x}_0).
    \label{eq:deterministic_duality}
\end{align}
Here, the second equality follows from integration by parts and the resulting adjoint relation between  $\mathcal{L}$ and $\mathcal{L}^{*}$ under suitable boundary conditions, and the last equality follows from the defining property of the Dirac delta distribution.

The adjoint operator $\mathcal{L}$ acts on observables and is the infinitesimal generator of the Koopman semigroup. For a sufficiently smooth observable $\phi$, it is given by
\begin{equation}
    \mathcal{L}\phi(\bm{x})
    =\sum_{i=1}^{D}f_i(\bm{x})
    \frac{\partial\phi(\bm{x})}{\partial x_i}
    =\bm{f}(\bm{x})\cdot\nabla\phi(\bm{x}).
    \label{eq:deterministic_koopman_generator}
\end{equation}
Therefore, the continuous-time Koopman operator is
\begin{equation}
    (\mathcal{K}_t\phi)(\bm{x}_0)
    =\phi(\bm{x}(t))
    =\left(e^{t\mathcal{L}}\phi\right)(\bm{x}_0).
    \label{eq:continuous_koopman_operator}
\end{equation}
For the sampling interval $\Delta t$ used in the preceding subsections, the discrete-time Koopman operator is $\mathcal{K}=\mathcal{K}_{\Delta t}$.

To obtain a finite-dimensional approximation of the Koopman operator,
we use the dictionary $\bm{\Psi}$ defined in Eq.~\eqref{eq:dictionary}. In this study, its $N_\mathrm{dic}$ dictionary functions are selected from the monomial basis functions
\begin{equation}
    \psi_{\bm{n}}(\bm{x})
    =
    x_1^{n_1}x_2^{n_2}\ldots x_D^{n_D},
    \label{eq:monomial_basis}
\end{equation}
where
$\bm{n}=(n_1,\ldots,n_D)\in\mathbb{N}_0^D$
is a multi-index.

For each $j=1,\ldots,N_{\mathrm{dic}}$, let $\phi_j(\bm{x},t)=(\mathcal{K}_t\psi_j)(\bm{x})$ denote the time evolution of the $j$-th dictionary function. For a fixed $j$, we suppress the $j$-dependence of the expansion coefficients and approximate $\phi_j$ as follows:
\begin{equation}
    \phi_j(\bm{x},t)
    \simeq
    \sum_{\bm{n}\in\mathcal{I}_{\mathrm{dic}}}c_{\bm{n}}(t)
    \psi_{\bm{n}}(\bm{x}),
    \label{eq:time_evolved_dictionary_expansion}
\end{equation}
where $\mathcal{I}_{\mathrm{dic}}\subset\mathbb{N}_0^D$ is a finite index set and $c_{\bm{n}}(t)$ are the corresponding time-dependent expansion coefficients. Although a finite truncation is required in practice, Eq.~\eqref{eq:time_evolved_dictionary_expansion} expresses the time-evolved dictionary function as a linear combination of monomials. Therefore, it directly corresponds to the finite-dimensional Koopman relation in Eq.~\eqref{eq:koopman on dictionary}.
The explicit derivation and numerical evaluation of the expansion coefficients are given in Appendix~\ref{app:coefficient_calculation}.

Although we use monomial basis functions in this study, other choices,
such as Hermite polynomials and radial basis functions, are also
possible. For non-monomial bases, however, the computation of the
finite-dimensional generator matrix is generally more involved.
Hybrid dictionaries combining monomials and radial basis functions
were discussed in Ref.~\cite{takahashi2023redundant}.
In this study, we restrict attention to monomial bases for simplicity.

\section{Proposed Method}
\label{sec:proposed_method}
\begin{figure}
    \centering
    \includegraphics[width=0.8\linewidth]{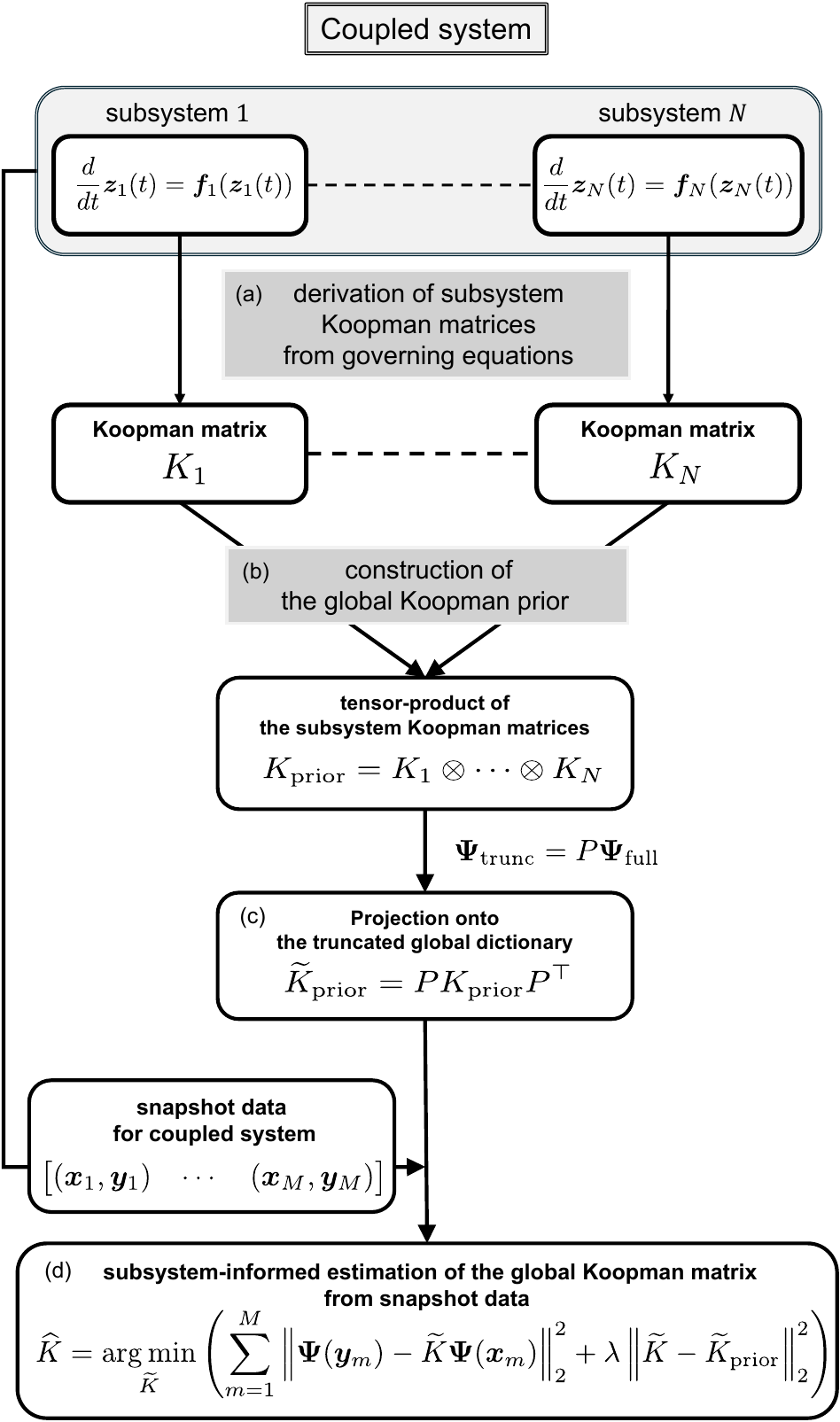}
    \caption{(Color online) Schematic overview of the proposed method. First, subsystem Koopman matrices are derived from the governing equations. Second, a tensor-product prior is constructed. Third, the prior is projected onto the truncated global dictionary. Finally, the global Koopman matrix is estimated from snapshot data using subsystem-informed EDMD.}
    \label{fig:overview}
\end{figure}
We describe the proposed method for constructing a finite-dimensional Koopman approximation for coupled systems using the differential equations governing each subsystem. Figure~\ref{fig:overview} provides a schematic overview.

The proposed method consists of four main steps. First, we derive Koopman matrix approximations for the individual subsystems from their governing differential equations using the method described in Sect.~\ref{sec:koopman_matrix_derivation_from_differential_equations}. Second, we construct a tensor-product prior for the global Koopman matrix. Third, the prior is projected onto the truncated global dictionary. Finally, the prior is incorporated as the center of a regularized EDMD problem, through which the coupling-induced correction is learned from snapshot data of the full coupled system.

To derive the subsystem Koopman matrices, we replace $\bm{f}$ in Eq.~\eqref{eq:continuous_dynamical_system} with the intrinsic dynamics $\bm{f}_i$ in Eq.~\eqref{eq:coupled_system} for each subsystem. We denote the Koopman matrix for the $i$-th subsystem at sampling interval $\Delta t$ by $K_i$. The derivation of the subsystem Koopman matrices is performed independently for each subsystem and does not require any information about the coupling functions $\bm{g}_{ij}$ or the coupling strengths $c_{ij}$.
In the following subsections, we describe the second through fourth steps in detail.

\subsection{Construction of tensor-product prior}
\label{sec:tensor_product_prior}
In this subsection, we construct a tensor-product prior for the global
Koopman matrix from the subsystem Koopman matrices. We first consider
the corresponding construction at the generator level.

Let $\mathcal{L}_i$ denote the Koopman generator associated with the $i$-th subsystem, derived from its intrinsic dynamics $\bm{f}_i$. If the subsystems are uncoupled, the Koopman generator of the full uncoupled system is given by
\begin{equation}
    \mathcal{L}_{\mathrm{prior}}
    =
    \sum_{i=1}^{N}
    I_1\otimes\ldots\otimes I_{i-1}
    \otimes \mathcal{L}_i
    \otimes I_{i+1}\otimes\ldots\otimes I_N,
    \label{eq:global_koopman_generator}
\end{equation}
where $I_i$ denotes the identity operator on the observable space of the $i$-th subsystem. We use this uncoupled-system generator as prior information for the coupled system.

In a finite-dimensional basis, Eq.~\eqref{eq:global_koopman_generator} is represented by the Kronecker sum
\begin{equation}
    L_{\mathrm{prior}}
    =
    L_1\oplus L_2\oplus\ldots\oplus L_N,
    \label{eq:global_koopman_generator_matrix}
\end{equation}
where $L_i$ is the finite-dimensional generator matrix of the $i$-th subsystem. Thus, $L_{\mathrm{prior}}$ is a finite-dimensional approximation of $\mathcal{L}_{\mathrm{prior}}$.

For a sampling interval $\Delta t$, the corresponding prior Koopman matrix is
\begin{align}
    K_{\mathrm{prior}}
    &= \exp\left(\Delta t L_{\mathrm{prior}}\right) \notag\\
    &= \exp\left[
        \Delta t
        \left(
            L_1\oplus L_2\oplus\ldots\oplus L_N
        \right)
    \right] \notag\\
    &= \exp\left(\Delta t L_1\right)
       \otimes
       \exp\left(\Delta t L_2\right)\otimes\ldots\otimes
       \exp\left(\Delta t L_N\right) \notag\\
    &= K_1\otimes K_2\otimes\ldots\otimes K_N.
    \label{eq:global_koopman_matrix}
\end{align}

\subsection{Projection onto a truncated global dictionary}
\label{sec:projection_truncated_dictionary}
However, the dimension of the full tensor-product dictionary can become
prohibitively large as the number of subsystems increases. Let
\begin{equation}
    \bm{\Psi}_i(\bm{x}_i)
    =
    \begin{bmatrix}
        \psi_{i,1}(\bm{x}_i) &
        \ldots &
        \psi_{i,N_{\mathrm{dic},i}}(\bm{x}_i)
    \end{bmatrix}^{\top},
    \label{eq:subsystem_dictionary}
\end{equation}
denote the dictionary vector for the $i$-th subsystem. In this study,
$\bm{\Psi}_i$ consists of all monomials in the $D_i$ state variables
whose total degree does not exceed $d_i$. Its dimension is therefore
\begin{equation}
    N_{\mathrm{dic},i}
    =
    \binom{D_i+d_i}{d_i}.
    \label{eq:subsystem_dictionary_dimension}
\end{equation}

The full tensor-product dictionary for the full uncoupled system is
defined as
\begin{equation}
    \bm{\Psi}_{\mathrm{full}}(\bm{x})
    =
    \bm{\Psi}_1(\bm{x}_1)
    \otimes
    \bm{\Psi}_2(\bm{x}_2)
    \otimes\ldots\otimes
    \bm{\Psi}_N(\bm{x}_N),
    \label{eq:full_tensor_dictionary}
\end{equation}
where
$\bm{x}
=
[\bm{x}_1^{\top},\ldots,\bm{x}_N^{\top}]^{\top}$.
The dimension of this dictionary is
\begin{equation}
    N_{\mathrm{dic},\mathrm{prior}}
    =
    \prod_{i=1}^{N}N_{\mathrm{dic},i}
    =
    \prod_{i=1}^{N}
    \binom{D_i+d_i}{d_i}.
    \label{eq:full_tensor_dictionary_dimension}
\end{equation}
Thus, for subsystem dictionaries of fixed dimension, the dimension of
the full tensor-product dictionary grows exponentially with the number
of subsystems.

To reduce the dimension of the global observable space, we retain only
the monomials whose total degree does not exceed a prescribed degree
$d_{\mathrm{trunc}}$. Let
$\bm{\Psi}_{\mathrm{trunc}}(\bm{x})
\in\mathbb{R}^{N_{\mathrm{trunc}}}$
denote the resulting truncated global dictionary vector. It is obtained
from the full tensor-product dictionary as
\begin{equation}
    \bm{\Psi}_{\mathrm{trunc}}(\bm{x})
    =
    P\bm{\Psi}_{\mathrm{full}}(\bm{x}),
    \label{eq:truncated_dictionary_vector}
\end{equation}
where
$P\in\mathbb{R}^{N_{\mathrm{trunc}}\times N_{\mathrm{dic},\mathrm{prior}}}$
is a selection matrix whose rows are extracted from the identity matrix
$I_{N_{\mathrm{dic},\mathrm{prior}}}$. Specifically, the truncated dictionary is indexed by
\begin{equation}
    \mathcal{I}_{\mathrm{trunc}}
    =
    \left\{
        \bm{n}\in\mathbb{N}_0^D
        \;\middle|\;
        |\bm{n}|_1\leq d_{\mathrm{trunc}}
    \right\},
    \label{eq:truncated_index_set}
\end{equation}
subject to the degree limits imposed by the subsystem dictionaries.
Accordingly,
\begin{equation}
    \bm{\Psi}_{\mathrm{trunc}}(\bm{x})
    =
    \left[
        \bm{x}^{\bm{n}}
    \right]_{\bm{n}\in\mathcal{I}_{\mathrm{trunc}}}.
    \label{eq:truncated_dictionary_expansion}
\end{equation}

The tensor-product Koopman matrix
$K_{\mathrm{prior}}=K_1\otimes\ldots\otimes K_N$
acts on the full dictionary in Eq.~\eqref{eq:full_tensor_dictionary}.
We define its compression to the truncated dictionary by
\begin{equation}
    \widetilde{K}_{\mathrm{prior}}
    =
    P K_{\mathrm{prior}}P^{\top}.
    \label{eq:truncated_prior}
\end{equation}
The matrix $\widetilde{K}_{\mathrm{prior}}$ is used as prior information
for estimating the Koopman matrix on the truncated global dictionary.

For the actual coupled system, the tensor-product prior derived in Eq.~\eqref{eq:global_koopman_matrix} does not generally coincide with the Koopman matrix of the coupled system because the prior is constructed under the assumption that the subsystems are uncoupled. Therefore, we estimate the Koopman matrix of the coupled system from snapshot data while incorporating the tensor-product prior. In this way, the known intrinsic dynamics of the subsystems are retained through the prior, whereas the effects of the unknown interactions are inferred from data.

\subsection{Subsystem-informed EDMD}
\label{sec:subsystem_informed_edmd}
In this subsection, we describe the subsystem-informed EDMD method for estimating the Koopman matrix of the coupled system from snapshot data. The method is based on the standard EDMD formulation, but it incorporates a regularization term that penalizes deviations from the truncated prior Koopman matrix $\widetilde{K}_{\mathrm{prior}}$.

Here, we denote the snapshot data of the coupled system $(\bm{z}(t_m),\bm{z}(t_{m+1}))$ by $(\bm{x}_m,\bm{y}_m)$ for $m=1,\ldots,M$.
Then, the subsystem-informed EDMD problem is formulated as follows:
\begin{widetext}
\begin{align}
    \widehat{K}
    =\operatorname*{arg\,min}_{\widetilde{K}}\Bigg\{
        \sum_{m=1}^{M}
        \left\|
            \bm{\Psi}_{\mathrm{trunc}}(\bm{y}_m)
            -
            \widetilde{K}\bm{\Psi}_{\mathrm{trunc}}(\bm{x}_m)
        \right\|_2^2
    + 
        \lambda
        \left\|
            \widetilde{K}
            -
            \widetilde{K}_{\mathrm{prior}}
        \right\|_F^2
    \Bigg\},
    \label{eq:prior_regularized_edmd}
\end{align}
\end{widetext}
where $\lambda>0$ is a regularization parameter that controls the trade-off between fitting the snapshot data and adhering to the prior. The first term in the cost function measures the difference between the predicted and actual observables, while the second term penalizes deviations from the prior Koopman matrix.

The minimizer of Eq.~\eqref{eq:prior_regularized_edmd} admits the
following closed-form solution:
\begin{widetext}
\begin{align}
    \widehat{K}=
    \left(
        \sum_{m=1}^{M}
        \bm{\Psi}_{\mathrm{trunc}}(\bm{y}_m)
        \bm{\Psi}_{\mathrm{trunc}}(\bm{x}_m)^{\top}
        +
        \lambda \widetilde{K}_{\mathrm{prior}}
    \right)
    \left(
        \sum_{m=1}^{M}
        \bm{\Psi}_{\mathrm{trunc}}(\bm{x}_m)
        \bm{\Psi}_{\mathrm{trunc}}(\bm{x}_m)^{\top}
        +
        \lambda I
    \right)^{-1}.
    \label{eq:closed_form_si_edmd}
\end{align}
\end{widetext}

The proposed method is related to the duality-based online EDMD method developed in Ref.~\cite{ohta2025integrated}. That method is based on an online EDMD framework, in which the Koopman matrix is updated sequentially as new snapshot data become available. Online EDMD is particularly useful for streaming data and systems whose dynamics may vary over time.

In the present study, however, we focus on a time-invariant system and assume that all snapshot data are available in advance. We therefore adopt a batch EDMD formulation, which provides a more direct implementation for the present setting and can efficiently exploit matrix operations on the complete dataset.

Under the time-invariant assumption, the same batch data can also be processed sequentially using online EDMD. The resulting online algorithm can be interpreted as a recursive implementation of the batch least-squares problem. Details of this relationship are provided in Appendix~\ref{app:online_edmd}.

The main distinction from Ref.~\cite{ohta2025integrated} is not the
choice between batch and online computation. Rather, the present method
constructs a tensor-product prior from the known intrinsic dynamics of
the subsystems and uses this prior to estimate the Koopman matrix of the
coupled system with unknown interactions.

In summary, the proposed method constructs a tensor-product prior from
the known subsystem dynamics and incorporates it into EDMD for the full
coupled system. This formulation preserves the available subsystem
information while allowing the effects of unknown interactions to be
identified from snapshot data.

\section{Numerical Experiments}
\label{sec:numerical_experiments}
In this study, data efficiency refers to achieving improved out-of-sample prediction accuracy with a limited number of snapshot pairs from the coupled system.
In this section, we present numerical experiments to demonstrate the effectiveness of the proposed method. We consider two coupled systems: coupled Duffing oscillators and coupled van der Pol oscillators. In each case, we compare the performance of the proposed method with standard EDMD and $\ell_2$-regularized EDMD. 

Standard EDMD serves as the unregularized baseline. The $\ell_2$-regularized EDMD baseline uses the same data-fitting loss as the proposed method, but penalizes $\|\widetilde K\|_F^2$ rather than $\|\widetilde K-\widetilde K_{\mathrm{prior}}\|_F^2$. Thus, comparing the two regularized methods at the same $\lambda$ examines the effect of centering the penalty at the subsystem-informed prior instead of zero. Note that, although $\ell_2$-regularized EDMD is generally employed to prevent overfitting, we include it here as a baseline for evaluating the effect of the regularization center.

All experiments are repeated for $20$ independent random seeds, and the results are summarized using the median and interquartile range.
For each simulated trajectory, the physical state is denoted by $\bm{z}(t)$, whereas the discrete snapshot data used in EDMD are denoted by
$\bm{x}_m=\bm{z}(t_m)$ and $\bm{y}_m=\bm{z}(t_{m+1})$.

To evaluate data efficiency, we vary the number of training snapshot pairs from $500$ to $5000$ in increments of $500$. For each training-data size, the Koopman matrices are independently computed using the proposed method, standard EDMD, and $\ell_2$-regularized EDMD, and their one-step prediction errors are compared as functions of the number of training snapshot pairs. The multi-step prediction performance is evaluated separately by fixing the number of training snapshot pairs at $1000$ and comparing the prediction errors over the prediction horizon.
\subsection{Coupled Duffing oscillators}
\label{sec:coupled_duffing_oscillators}
We consider the following coupled Duffing oscillators
\begin{align}
\dot{z}_{i,1}
&=z_{i,2}, \notag\\
\dot{z}_{i,2}
&=-\delta_i z_{i,2}
-\alpha_i z_{i,1}
-\beta_i z_{i,1}^{3} 
+\sum_{j\in\mathcal{N}_i}
c_{ij}(z_{j,1}-z_{i,1}),
\label{eq:coupled_duffing_oscillators}
\end{align}
where $i=1,2,3$, $\delta_{i}$, $\alpha_{i}$, and $\beta_{i}$ are parameters of the $i$-th oscillator, and $c_{ij}=1.0$ if $j\in\mathcal{N}_i$; otherwise $c_{ij}=0$. The neighbor sets are given by $\mathcal{N}_1=\{2\}$, $\mathcal{N}_2=\{1,3\}$, and $\mathcal{N}_3=\{2\}$. Hereafter, $\operatorname{Unif}(a,b)$ denotes the uniform distribution on the interval $(a,b)$, and $\operatorname{Unif}([a,b]^d)$ denotes the uniform distribution on the $d$-dimensional hypercube $[a,b]^d$. The parameters of the $i$-th oscillator are generated from uniform distributions as follows:
\begin{itemize}
    \item $\delta_{i}\sim \operatorname{Unif}(0.1,0.3)$,
    \item $\alpha_{i}\sim \operatorname{Unif}(-1.0,-0.5)$,
    \item $\beta_{i}\sim \operatorname{Unif}(0.5,1.0)$.
\end{itemize}
The training initial condition is sampled as
\begin{equation}
    \bm{z}^{\mathrm{train}}(0)
    \sim \operatorname{Unif}([-1.5,1.5]^6).
\end{equation}
We then generate $M_{\mathrm{train}}+1=5001$ state snapshots with a sampling interval $\Delta t=0.01$ by numerically integrating the above equations using the \texttt{solve\_ivp} function in SciPy.\cite{virtanen2020scipy} We use its default RK45 solver. These snapshots yield $M_{\mathrm{train}}=5000$ training snapshot pairs.

In this study, we intentionally consider a setting in which only a single training trajectory is available. As shown in Sect.~\ref{sec:coupled_van_der_pol_oscillators}, this trajectory gradually approaches an equilibrium point, so increasing its length does not provide substantial additional information about the transient dynamics. This example is used to examine whether the subsystem-informed prior can compensate for the limited dynamical information contained in the training data.

For the coupled Duffing system, the subsystem dictionaries and the truncated global dictionary are selected as monomials with $d_i=d_{\mathrm{trunc}}=3$, and the regularization parameter $\lambda$ is set to $1.0$ for both the proposed method and $\ell_2$-regularized EDMD. Note that we will discuss other $\lambda$ cases later.

For performance evaluation, we generate $N_{\mathrm{test}}=50$ test trajectories, each containing $M_{\mathrm{test}}+1=1001$ state snapshots and hence $M_{\mathrm{test}}=1000$ snapshot pairs. We assume that the physical initial state of each test trajectory is close to that of the training trajectory. Hence, the initial state of test trajectory $i$ is generated as follows:
\begin{equation}
    \bm{z}^{\mathrm{test},i}(0)
    =\bm{z}^{\mathrm{train}}(0)+\bm{\epsilon}^{(i)},
    \label{eq:duffing_test_initial_condition}
\end{equation}
where $\bm{\epsilon}^{(i)}\sim \operatorname{Unif}([-0.3,0.3]^6)$ is a small perturbation. The corresponding test snapshot data are denoted by $\bm{x}_m^{\mathrm{test},i}=\bm{z}^{\mathrm{test},i}(t_m)$.

To evaluate the prediction performance, we calculate the multi-step predictions as follows:
\begin{align}
\widehat{\bm{\Psi}}_{m+n}^{\mathrm{test},i}
&=\widehat{K}^{n}\bm{\Psi}_{\mathrm{trunc}}(\bm{x}_{m}^{\mathrm{test},i}), \notag\\
\widehat{\bm{x}}_{m+n}^{\mathrm{test},i}
&=B\widehat{\bm{\Psi}}_{m+n}^{\mathrm{test},i},
\label{eq:multi_step_state_prediction}
\end{align}
where $n$ is the prediction horizon, $\widehat{K}$ is the Koopman matrix estimated by each method, $\widehat{\bm{\Psi}}_{m+n}^{\mathrm{test},i}$ is the predicted lifted state, $\widehat{\bm{x}}_{m+n}^{\mathrm{test},i}$ is the corresponding state prediction, and $B\in\mathbb{R}^{D\times N_{\mathrm{trunc}}}$ is the projection matrix that maps the dictionary functions to the state space.
Note that, as denoted in Sects.~\ref{sec:preliminaries} and ~\ref{sec:proposed_method}, the vector-valued full-state observable $\bm{\phi}_{\mathrm{state}}(\bm{x})=\bm{x}$ is represented by $\bm{\Psi}_{\mathrm{trunc}}$, because each state component $x_r$ is included as a degree-one monomial. The matrix $B$ selects these components, i.e., $B\bm{\Psi}_{\mathrm{trunc}}(\bm{x})=\bm{x}$. Hence, Eq.~\eqref{eq:multi_step_state_prediction} evolves the dictionary and extracts the predicted state.

For the one-step prediction evaluation, we collect the true and predicted states of test trajectory $i$ into the data matrices
\begin{align}
    X^{\mathrm{test},i}
    &=
    \begin{bmatrix}
        \bm{x}_{1}^{\mathrm{test},i} & \ldots & \bm{x}_{M_{\mathrm{test}}}^{\mathrm{test},i}
    \end{bmatrix}, \notag\\
    \widehat{X}^{\mathrm{test},i}
    &=
    \begin{bmatrix}
        \widehat{\bm{x}}_{1}^{\mathrm{test},i} & \ldots & \widehat{\bm{x}}_{M_{\mathrm{test}}}^{\mathrm{test},i}
    \end{bmatrix}.
\label{eq:one_step_test_data_matrices}
\end{align}
The one-step prediction error for test trajectory $i$ is defined by the relative Frobenius norm
\begin{equation}
    e_{\mathrm{one-step}}^{(i)}
    =
    \frac{
        \left\|X^{\mathrm{test},i}-\widehat{X}^{\mathrm{test},i}\right\|_{F}
    }{
        \left\|X^{\mathrm{test},i}\right\|_{F}+10^{-10}
    }.
    \label{eq:one_step_prediction_error}
\end{equation}
Since this error is normalized by the Frobenius norm of the true state matrix, it is dimensionless and measures the prediction discrepancy relative to the magnitude of the state data, rather than relative to the one-step state increment.

To separately assess the accuracy of the predicted one-step state evolution, we also evaluate the prediction error normalized by the magnitude of the true one-step increment. The definition and numerical results of this alternative error metric are provided in Appendix~\ref{app:increment_error}.

\begin{figure}[t]
    \centering
    \includegraphics[width=\linewidth]{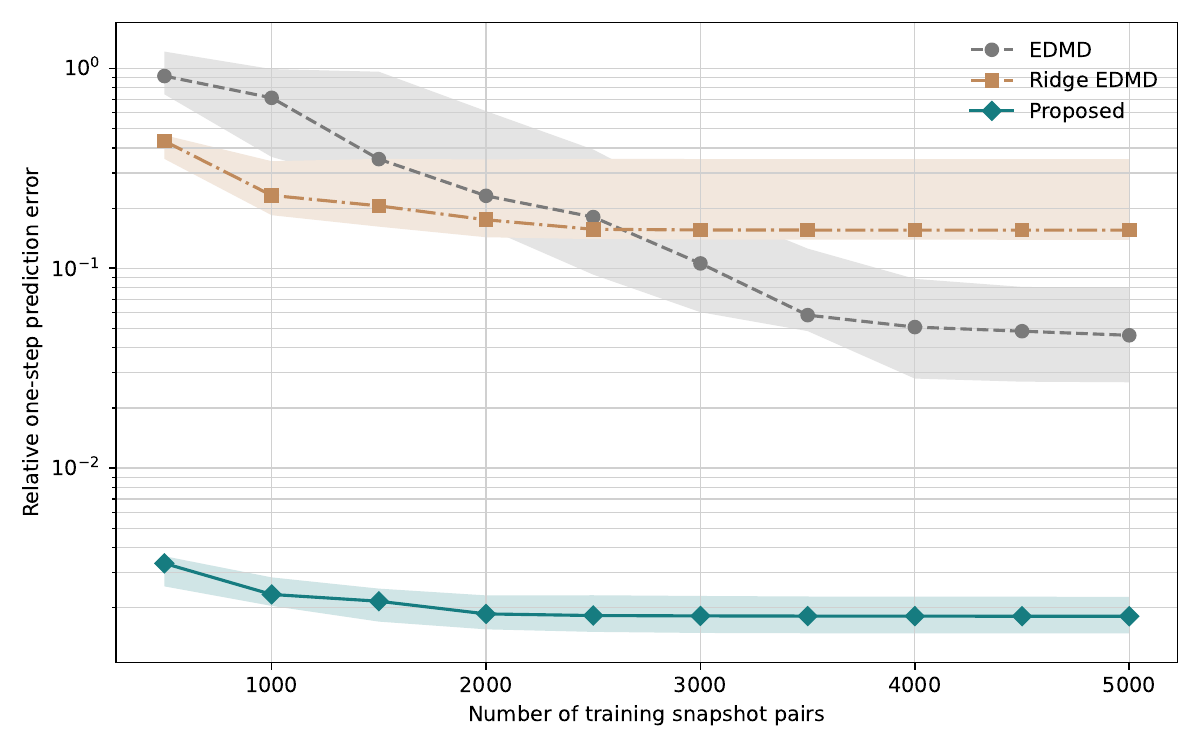}
    \caption{(Color online) The median and interquartile range of the one-step prediction error $\overline{e}^{\mathrm{one-step}}$ across $20$ independent random seeds for the coupled Duffing system. The horizontal axis represents the number of training snapshot pairs, and the vertical axis represents the one-step prediction error. 
    The line and shaded area represent the median and interquartile range, respectively. }
    \label{fig:duffing_one_step}
\end{figure}

First, we evaluate the one-step prediction error for each method. For each random seed, we average the errors over all test trajectories as
\begin{equation}
    \overline{e}^{\mathrm{one-step}}
    =
    \frac{1}{N_{\mathrm{test}}}
    \sum_{i=1}^{N_{\mathrm{test}}}
    e_{\mathrm{one-step}}^{(i)}.
    \label{eq:mean_one_step_prediction_error}
\end{equation}
Then, we summarize the median and interquartile range of $\overline{e}^{\mathrm{one-step}}$ across $20$ independent random seeds. Figure~\ref{fig:duffing_one_step} shows the results of the one-step prediction error for each method. 
From Fig.~\ref{fig:duffing_one_step}, we see that the proposed method yields a lower one-step prediction error than the other methods, especially when the number of training snapshot pairs is small. Moreover, the proposed method exhibits a lower prediction error than $\ell_2$-regularized EDMD. 
At the same regularization parameter, the proposed method yields a lower prediction error than zero-centered $\ell_2$-regularized EDMD. Since the two methods use the same form of quadratic regularization and differ in its center, this result indicates that the subsystem-informed prior provides a more informative regularization center than the zero matrix.

Next, we evaluate the multi-step prediction error for each method. Starting from the initial data point $\bm{x}_0^{\mathrm{test},i}$ of each test trajectory, we compute the prediction error at each future time step $n=1,2,\ldots,100$. In calculating the multi-step prediction error, we use the Koopman matrix estimated from $1000$ training snapshot pairs for each method.
For test trajectory $i$, the relative multi-step prediction error at prediction step $n$ is defined as
\begin{equation}
    e_{n}^{(i)}
    =
    \frac{
        \left\|\bm{x}_{n}^{\mathrm{test},i}
        -\widehat{\bm{x}}_{n}^{\mathrm{test},i}\right\|_2
    }{
        \left\|\bm{x}_{n}^{\mathrm{test},i}\right\|_2+10^{-10}
    }.
    \label{eq:multi_step_prediction_error}
\end{equation}
For each random seed, we average the multi-step prediction error over all test trajectories as
\begin{equation}
    \overline{e}^{\mathrm{multi-step}}_{n}
    =
    \frac{1}{N_{\mathrm{test}}}
    \sum_{i=1}^{N_{\mathrm{test}}}
    e_{n}^{(i)}.
    \label{eq:mean_multi_step_prediction_error}
\end{equation}
The one-step evaluation uses all snapshot pairs to assess prediction accuracy along each test trajectory, whereas the multi-step evaluation starts from $m=0$ to examine the accumulation of recursively propagated prediction errors over a fixed prediction horizon.

\begin{figure}[t]
    \centering
    \includegraphics[width=\linewidth]{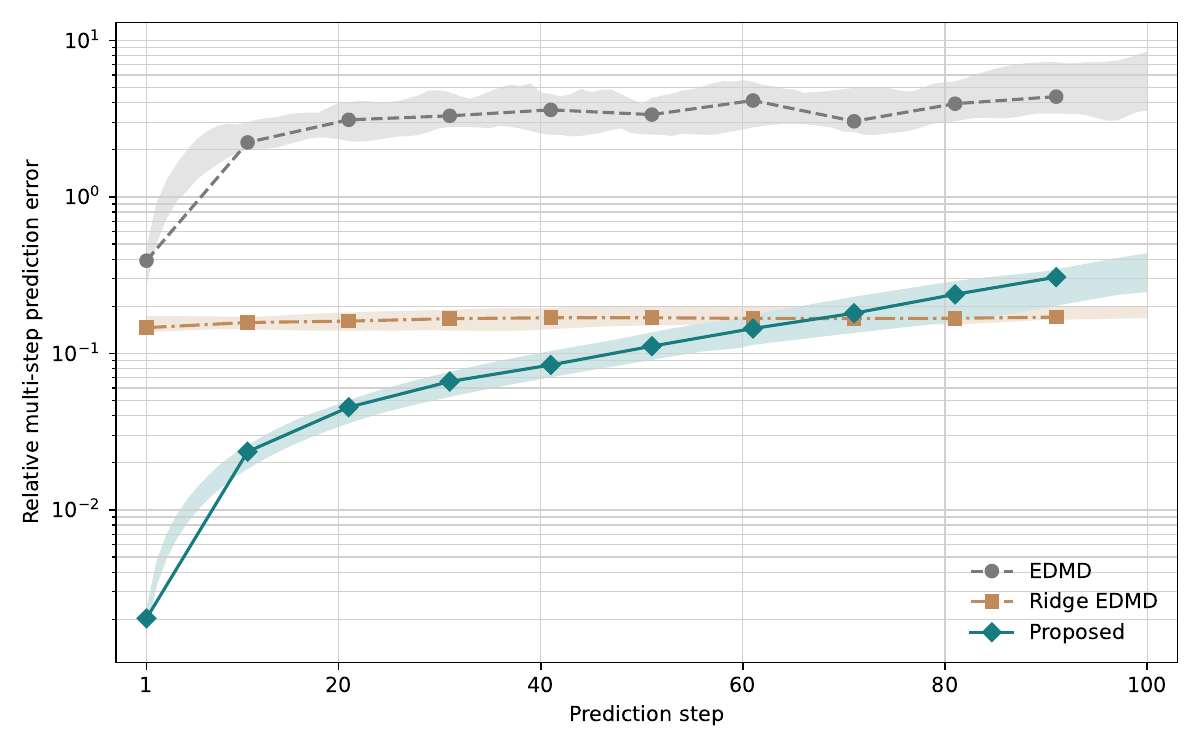}
    \caption{(Color online) The median and interquartile range of the multi-step prediction error $\overline{e}^{\mathrm{multi-step}}_{n}$ across $20$ independent random seeds for the coupled Duffing system, using Koopman matrices estimated from $1000$ training snapshot pairs. The horizontal axis represents the prediction horizon, and the vertical axis represents the multi-step prediction error. The line and shaded area represent the median and interquartile range, respectively.}
    \label{fig:duffing_multi_step}
\end{figure}

Then, we summarize the median and interquartile range of $\overline{e}^{\mathrm{multi-step}}_{n}$ across $20$ independent random seeds. Figure~\ref{fig:duffing_multi_step} shows the results of the multi-step prediction error for each method.

From Fig.~\ref{fig:duffing_multi_step}, we see that the proposed method achieves lower multi-step prediction errors than EDMD, especially for short-term predictions. However, the prediction error of the proposed method gradually increases as the prediction horizon becomes longer and exceeds that of $\ell_2$-regularized EDMD after approximately $n=70$. This indicates that the proposed method effectively incorporates the prior information from the subsystem dynamics, improving the accuracy of the estimated Koopman matrix for short-term prediction. However, the prior information alone does not necessarily improve long-term prediction robustness because the accumulation of prediction errors dominates the dynamics.

Although the proposed method maintains a lower multi-step prediction error than EDMD, the errors of both methods increase as the prediction horizon grows. This shared upward trend suggests that incorporating subsystem prior information improves short-term prediction accuracy but does not, by itself, prevent error growth over longer prediction horizons.

The relatively weak growth of the multi-step prediction error of $\ell_2$-regularized EDMD may be attributed to the effect of regularization, which reduces the spectral radius of the estimated Koopman matrix compared with standard EDMD and the proposed method. Consequently, the amplification of Koopman modes during repeated prediction is suppressed. However, this weak error growth does not necessarily indicate accurate long-term prediction because its one-step prediction error remains relatively large.

\begin{figure}[t]
    \centering
    \includegraphics[width=\linewidth]{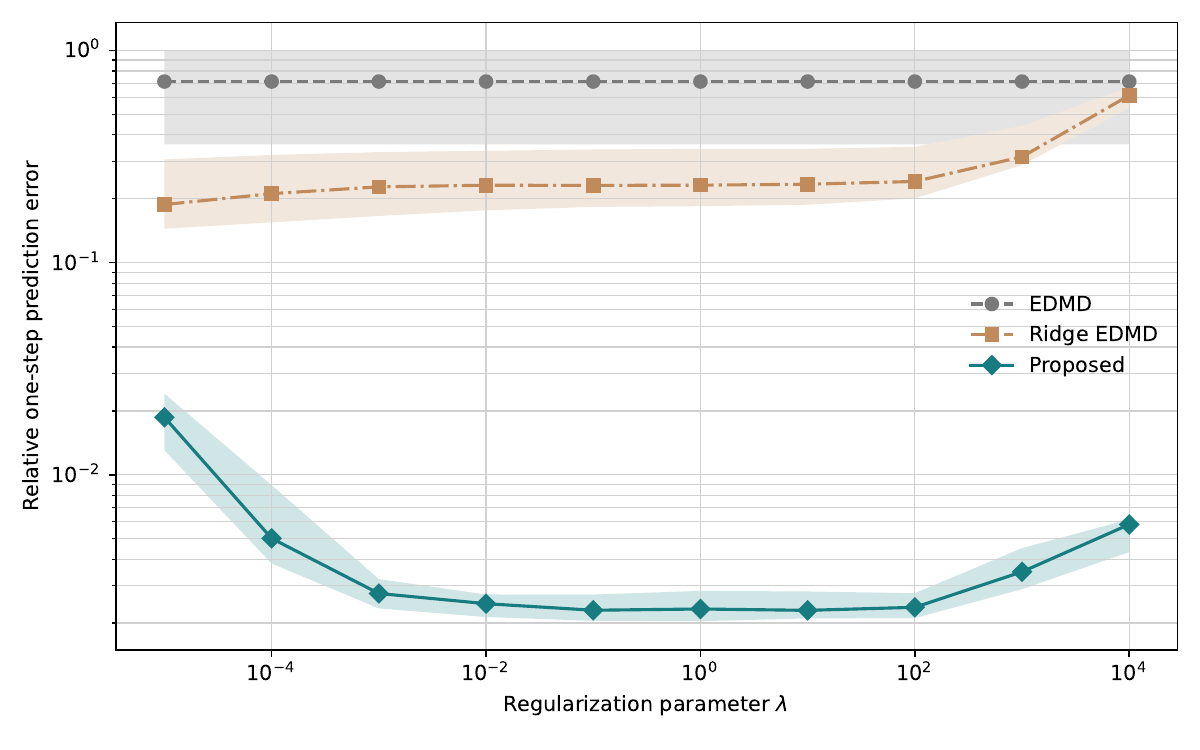}
    \caption{(Color online) Effects of the regularization parameter $\lambda$ on the one-step prediction error $\overline{e}^{\mathrm{one-step}}$ across $20$ independent random seeds for the coupled Duffing system. The horizontal axis represents the regularization parameter $\lambda$, and the vertical axis represents the one-step prediction error. The line and shaded area represent the median and interquartile range, respectively. The standard EDMD curve is horizontal because standard EDMD does not use the regularization parameter $\lambda$; it is shown as a $\lambda$-independent reference.}
    \label{fig:duffing_lambda_sensitivity}
\end{figure}

Finally, we investigate the sensitivity of the proposed method to the regularization parameter $\lambda$. We vary $\lambda$ from $10^{-5}$ to $10^{4}$ and evaluate the one-step prediction error for each value. Figure~\ref{fig:duffing_lambda_sensitivity} shows the results of the sensitivity analysis. Here, we use the training dataset of $1000$ snapshot pairs for each method.

From Fig.~\ref{fig:duffing_lambda_sensitivity}, we see that, at $\lambda=10^{-5}$, the proposed method is almost equivalent to $\ell_2$-regularized EDMD, and the one-step prediction error is relatively high. As $\lambda$ increases, the one-step prediction error of the proposed method decreases and remains nearly constant over the range $10^{-1}\leq\lambda\leq10^1$, with its minimum attained at approximately $\lambda=10^1$, whereas that of $\ell_2$-regularized EDMD gradually increases. This indicates that the proposed method effectively incorporates the prior information from the subsystem dynamics, improving the accuracy of the estimated Koopman matrix.

However, as $\lambda$ increases further, the one-step prediction errors for the proposed method gradually increase. This is because a large $\lambda$ forces the estimated Koopman matrix to be close to the prior, which may not accurately capture the dynamics of the coupled system, especially the interaction terms. Therefore, the choice of $\lambda$ is crucial for balancing the influence of the prior and the data, and it should be selected based on the specific characteristics of the system and the available data.

\subsection{Coupled van der Pol oscillators}
\label{sec:coupled_van_der_pol_oscillators}
Next, we consider the following coupled van der Pol oscillators
\begin{align}
\dot{z}_{i,1}
&=z_{i,2}, \notag\\
\dot{z}_{i,2}
&=\mu_i(1-z_{i,1}^2)z_{i,2}-z_{i,1}+\sum_{j\in\mathcal{N}_i}c_{ij}(z_{j,1}-z_{i,1}),
\label{eq:coupled_van_der_pol_oscillators}
\end{align}
where $i=1,2,3$, $\mu_{i}$ is a parameter of the $i$-th oscillator, and $c_{ij}=1.0$ if $j\in\mathcal{N}_i$; otherwise $c_{ij}=0$. The neighbor sets are given by $\mathcal{N}_1=\{2\}$, $\mathcal{N}_2=\{1,3\}$, and $\mathcal{N}_3=\{2\}$. The parameter $\mu_{i}$ is independently sampled from $\operatorname{Unif}(1.0,1.5)$ for each oscillator. The training initial condition is sampled as
\begin{equation}
    \bm{z}^{\mathrm{train}}(0)
    \sim
    \left[
        \operatorname{Unif}\left(-\frac{\pi}{2},\frac{\pi}{2}\right)
        \times \operatorname{Unif}(-1,1)
    \right]^{\otimes 3}.
\end{equation}
We generate $6001$ state snapshots with a sampling interval $\Delta t=0.01$ by numerically integrating the above equations using SciPy's \texttt{solve\_ivp} function with the RK45 solver.\cite{virtanen2020scipy,dormand1980family} We then discard the first $1000$ snapshots as an initial transient and use the remaining $M_{\mathrm{train}}+1=5001$ snapshots, which yield $M_{\mathrm{train}}=5000$ training snapshot pairs. We use the same dictionary functions and regularization parameter $\lambda$ as in the coupled Duffing system.

For performance evaluation, we generate $N_{\mathrm{test}}=50$ test trajectories, each containing $3001$ state snapshots, and discard the first $1000$ snapshots as an initial transient. Thus, each retained test trajectory contains $M_{\mathrm{test}}+1=2001$ snapshots and yields $M_{\mathrm{test}}=2000$ snapshot pairs. The physical initial state of test trajectory $i$ is generated as follows:
\begin{equation}
    \bm{z}^{\mathrm{test},i}(0)
    =\bm{z}^{\mathrm{train}}(0)+\bm{\epsilon}_{\mathrm{vdp}}^{(i)},
    \label{eq:vdp_test_initial_condition}
\end{equation}
where $\bm{\epsilon}_{\mathrm{vdp}}^{(i)}\sim \operatorname{Unif}([-0.2,0.2]^6)$ is a small perturbation. The retained test snapshot data are denoted by $\bm{x}_m^{\mathrm{test},i}$.

For the coupled van der Pol system, we evaluate the one-step and multi-step prediction errors in the same manner as in the coupled Duffing system. 
\begin{figure}[t]
    \centering
    \includegraphics[width=\linewidth]{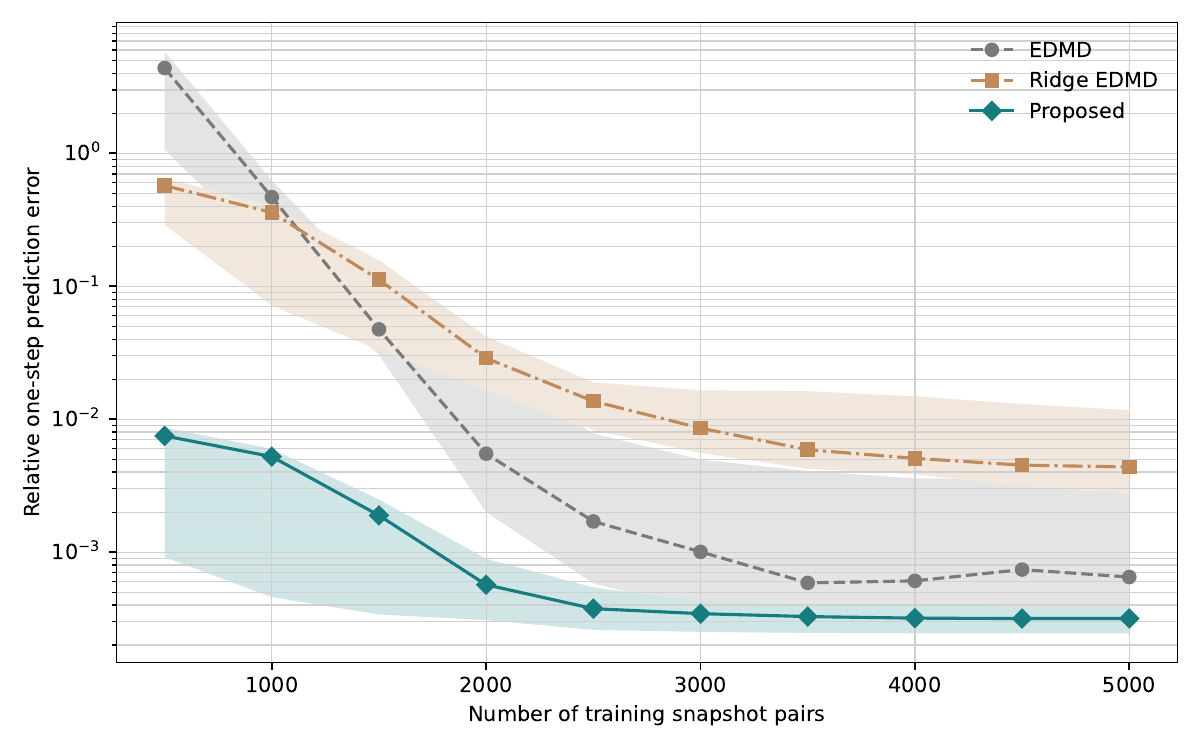}
    \caption{(Color online) The median and interquartile range of the one-step prediction error $\overline{e}^{\mathrm{one-step}}$ across $20$ independent random seeds for the coupled van der Pol system. The horizontal axis represents the number of training snapshot pairs, and the vertical axis represents the one-step prediction error. The line and shaded area represent the median and interquartile range, respectively.}
    \label{fig:vdp_one_step}
\end{figure}

Figure~\ref{fig:vdp_one_step} shows the results of the one-step prediction error for each method.
From Fig.~\ref{fig:vdp_one_step}, we see that the proposed method yields better performance than the other methods, especially when the number of training snapshot pairs is small.
At the same regularization parameter, the proposed method yields a lower prediction error than zero-centered $\ell_2$-regularized EDMD. This indicates that the subsystem-informed prior provides a more informative regularization center than the zero matrix.
When the number of training snapshot pairs is large, the proposed method is almost equivalent to EDMD. This is because the training dataset is sufficiently large to accurately estimate the Koopman matrix of the coupled system, and the influence of the prior information becomes less significant.

To examine the difference between the one-step prediction results for the two systems, we show the training trajectories.
\begin{figure}[t]
    \centering
    \includegraphics[width=\linewidth]{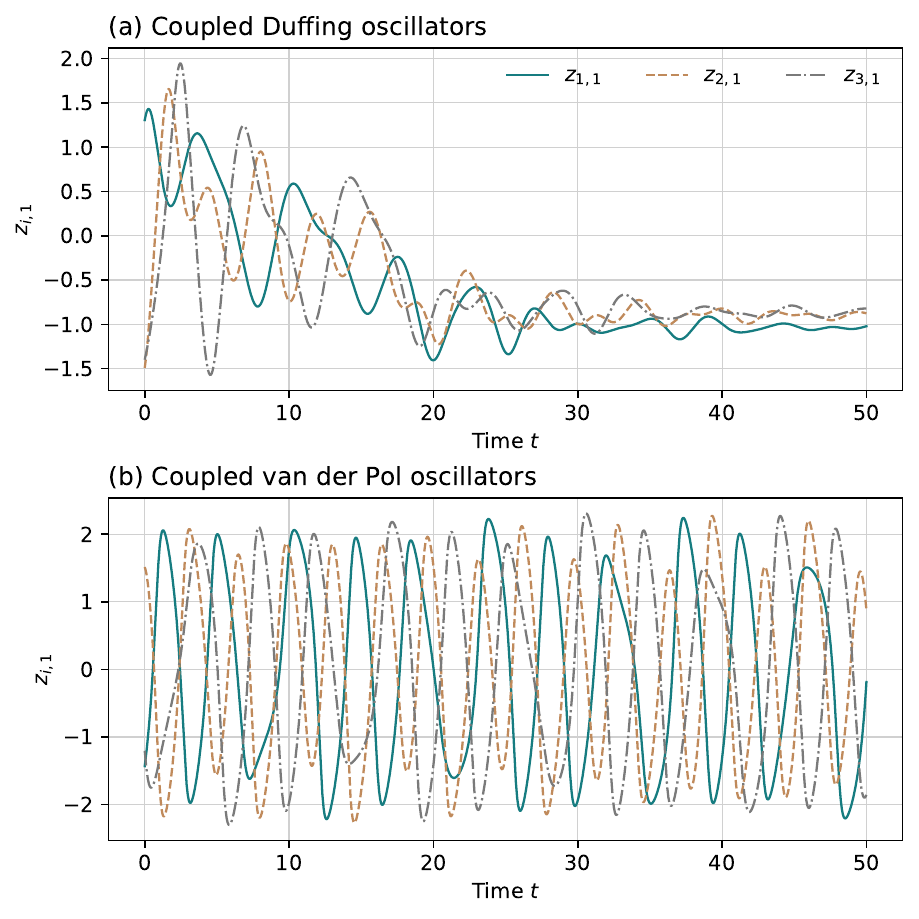}
    \caption{(Color online) Representative training trajectories for one
    random seed. The position components $z_{i,1}$ are shown for
    (a) the coupled Duffing system and (b) the coupled van der Pol system.
    For the coupled van der Pol system, the trajectories after removal of the
    initial transient are shown.}
    \label{fig:training_time_series}
\end{figure}
As shown in Fig.~\ref{fig:training_time_series}, the Duffing trajectories approach equilibrium states. Consequently, additional snapshot pairs collected at later times are concentrated near the equilibria and provide limited new information about the transient dynamics. This behavior is consistent with the saturation of the EDMD prediction error in Fig.~\ref{fig:duffing_one_step}. In contrast, the van der Pol trajectories continue to exhibit oscillatory behavior after the initial transient, as shown in Fig.~\ref{fig:training_time_series}(b). Therefore, the additional snapshot pairs continue to cover the oscillatory dynamics, which may explain why the EDMD error approaches that of the proposed method in Fig.~\ref{fig:vdp_one_step}.

\begin{figure}[t]
    \centering
    \includegraphics[width=\linewidth]{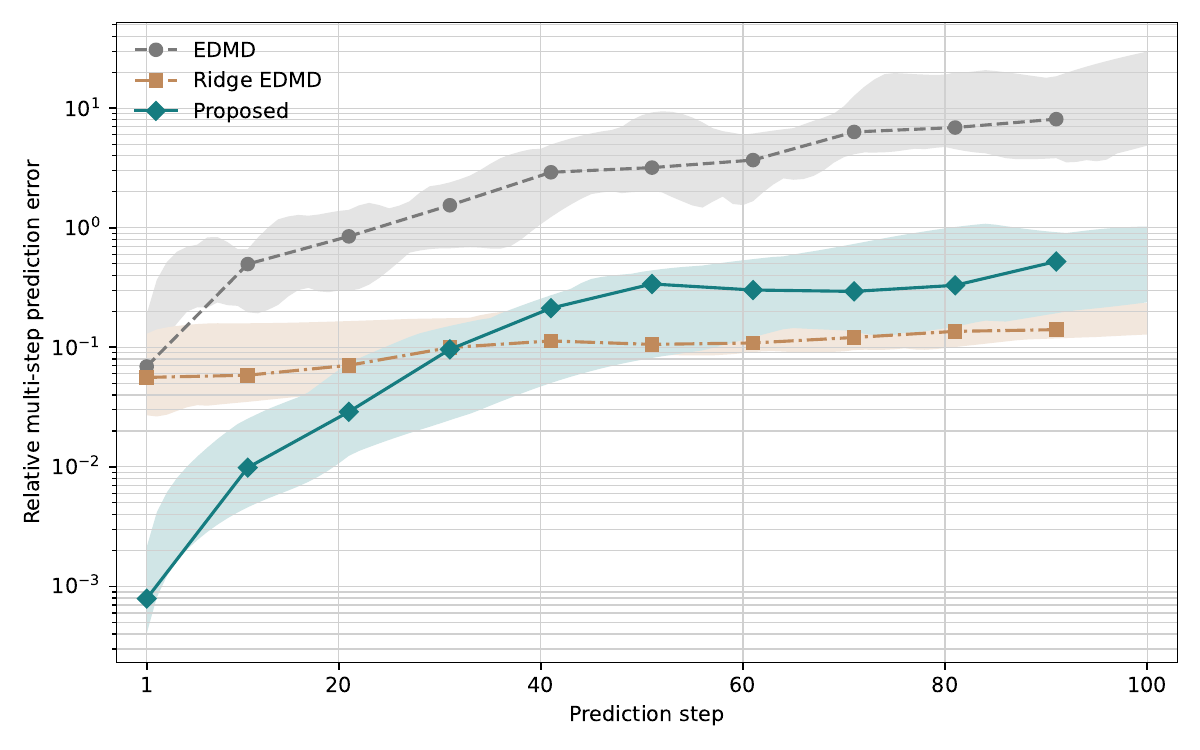}
    \caption{(Color online) The median and interquartile range of the multi-step prediction error $\overline{e}^{\mathrm{multi-step}}_{n}$ across $20$ independent random seeds for the coupled van der Pol system, using Koopman matrices estimated from $1000$ training snapshot pairs. The horizontal axis represents the prediction horizon, and the vertical axis represents the multi-step prediction error. The line and shaded area represent the median and interquartile range, respectively.}
    \label{fig:vdp_multi_step}
\end{figure}

Next, we evaluate the multi-step prediction error for each method in the same manner as in the coupled Duffing system, using the Koopman matrices estimated from $1000$ training snapshot pairs. Figure~\ref{fig:vdp_multi_step} shows the results of the multi-step prediction error for each method.
From Fig.~\ref{fig:vdp_multi_step}, we obtain almost the same results as for the coupled Duffing system. The proposed method achieves lower multi-step prediction errors than EDMD, especially for short-term predictions. However, the prediction error of the proposed method gradually increases as the prediction horizon becomes longer and exceeds that of $\ell_2$-regularized EDMD after approximately $n=30$. This leads to the same conclusion as for the coupled Duffing system: the proposed method effectively incorporates prior information from the subsystem dynamics, thereby improving the accuracy of the estimated Koopman matrix for short-term prediction. However, the prior information alone does not necessarily improve long-term prediction robustness because the accumulation of prediction errors dominates the dynamics.

Although the proposed method maintains a lower multi-step prediction error than EDMD, the errors of both methods increase as the prediction horizon grows. This shared upward trend suggests that incorporating subsystem prior information improves short-term prediction accuracy but does not, by itself, prevent error growth over longer prediction horizons.

As in the coupled Duffing system, $\ell_2$ regularization reduces the spectral radius of the estimated Koopman matrix compared with standard EDMD and the proposed method. Consequently, the amplification of Koopman modes during repeated prediction is suppressed, resulting in weaker growth of the multi-step prediction error. This behavior does not necessarily indicate accurate long-term prediction because its one-step prediction error remains relatively large.

\begin{figure}[t]
    \centering
    \includegraphics[width=\linewidth]{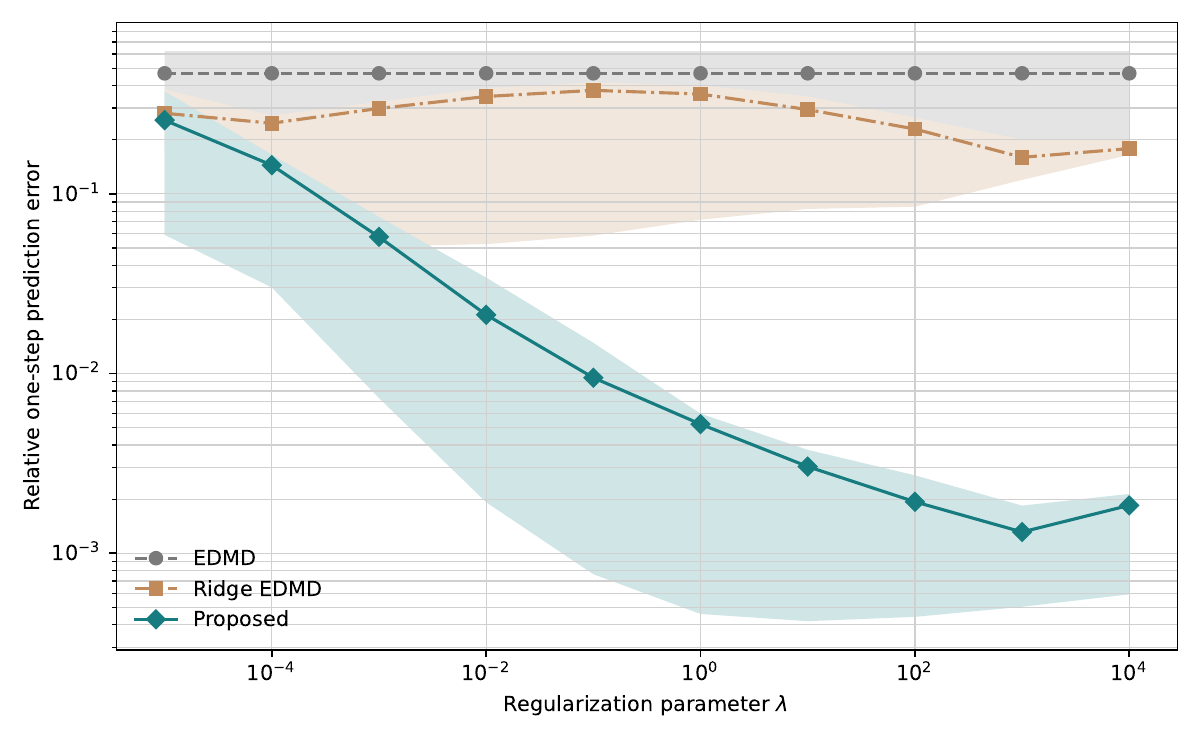}
    \caption{(Color online) Effects of the regularization parameter $\lambda$ on the one-step prediction error $\overline{e}^{\mathrm{one-step}}$ across $20$ independent random seeds for the coupled van der Pol system. The horizontal axis represents the regularization parameter $\lambda$, and the vertical axis represents the one-step prediction error. The line and shaded area represent the median and interquartile range, respectively. The standard EDMD curve is horizontal because standard EDMD does not use the regularization parameter $\lambda$; it is shown as a $\lambda$-independent reference.}
    \label{fig:vdp_lambda_sensitivity}
\end{figure}

Finally, we investigate the sensitivity of the proposed method to the regularization parameter $\lambda$ in the same manner as in the coupled Duffing system. Figure~\ref{fig:vdp_lambda_sensitivity} shows the results of the sensitivity analysis.
From Fig.~\ref{fig:vdp_lambda_sensitivity}, we see that, at $\lambda=10^{-5}$, the proposed method is almost equivalent to $\ell_2$-regularized EDMD. However, as $\lambda$ increases, the one-step prediction errors for the proposed method decrease and reach a minimum around $\lambda=10^{3}$, whereas those for $\ell_2$-regularized EDMD gradually increase. This indicates that the proposed method effectively incorporates the prior information from the subsystem dynamics, improving the accuracy of the estimated Koopman matrix.

\section{Conclusion}
\label{sec:conclusion}
In this study, we proposed subsystem-informed EDMD for learning the Koopman matrix of a coupled system by combining prior information from the known intrinsic dynamics of the subsystems with snapshot data of the full coupled system. Under the assumption that the subsystems are uncoupled, the subsystem Koopman matrices were combined through a tensor product to construct a prior for the global Koopman matrix. To avoid the rapid growth in the size of the tensor-product dictionary, the prior was restricted to a truncated global monomial dictionary. The Koopman matrix of the coupled system was then estimated by solving a regularized EDMD problem in which the tensor-product prior was incorporated.

Numerical experiments demonstrated that the proposed method improves the prediction accuracy compared with standard EDMD, particularly when the amount of training data is limited. The comparison with
$\ell_2$-regularized EDMD indicated that the improvement is attributable not merely to quadratic regularization, but specifically to regularization toward the subsystem-informed prior. The results also showed that the regularization parameter plays an important role in balancing the subsystem prior and the information contained in the snapshot data.

The advantage of the proposed method became less pronounced for long prediction horizons, suggesting that the subsystem prior improves the accuracy of the finite-dimensional Koopman approximation but does not by itself guarantee long-term prediction stability. Future work will focus on systematic selection of the regularization parameter, incorporation of stability-aware constraints, and efficient extensions to large-scale coupled systems.

\section*{Acknowledgments}
This work was financially supported in part by the JST FOREST Program (Grant Number JPMJFR216K, Japan) and JSPS KAKENHI Grant Number 26KJ0670.
\appendix
\section{Calculation of Expansion Coefficients}
\label{app:coefficient_calculation}
Here, we demonstrate how to calculate the expansion coefficients $c_{\bm{n}}(t)$ in Eq.~\eqref{eq:time_evolved_dictionary_expansion}.
For example, consider the Duffing oscillator
\begin{equation}
    \dot{x}_1=x_2,
    \qquad
    \dot{x}_2=-\delta x_2-\alpha x_1-\beta x_1^3,
    \label{eq:appendix_duffing}
\end{equation}
where $\delta$, $\alpha$, and $\beta$ are parameters. From Eq.~\eqref{eq:deterministic_koopman_generator}, the deterministic Koopman generator $\mathcal{L}$ is
\begin{equation}
    \mathcal{L}
    =x_2\frac{\partial}{\partial x_1}
    +\left(-\delta x_2-\alpha x_1-\beta x_1^3\right)
    \frac{\partial}{\partial x_2}.
    \label{eq:appendix_duffing_generator}
\end{equation}
First, we denote the observable $\phi(\bm{x}(t))$ by $\phi(\bm{x},t)$. From Eq.~\eqref{eq:continuous_koopman_operator}, the time evolution of $\phi(\bm{x},t)$ is governed by the backward equation
\begin{equation}
    \frac{\partial}{\partial t}\phi(\bm{x},t)
    =\mathcal{L}\phi(\bm{x},t).
    \label{eq:appendix_backward_equation}
\end{equation}
Substituting Eq.~\eqref{eq:appendix_duffing_generator} into Eq.~\eqref{eq:appendix_backward_equation}, we obtain
\begin{align}
    \frac{\partial}{\partial t}\phi(\bm{x},t)
    &=x_2\frac{\partial}{\partial x_1}\phi(\bm{x},t)\notag\\
    &\quad+\left(-\delta x_2-\alpha x_1-\beta x_1^3\right)
    \frac{\partial}{\partial x_2}\phi(\bm{x},t).
    \label{eq:appendix_backward_equation_explicit}
\end{align}
Next, substituting Eq.~\eqref{eq:time_evolved_dictionary_expansion} into Eq.~\eqref{eq:appendix_backward_equation_explicit}, we obtain

\begin{align}
\label{eq:appendix_backward_equation_expansion}
\frac{\partial}{\partial t}&\sum_{\bm{n}}c_{\bm{n}}(t)\psi_{\bm{n}}(\bm{x}) \notag\\
&=x_2\frac{\partial}{\partial x_1}\sum_{\bm{n}}c_{\bm{n}}(t)\psi_{\bm{n}}(\bm{x})\notag\\
&\quad+\left(-\delta x_2-\alpha x_1-\beta x_1^3\right)
\frac{\partial}{\partial x_2}\sum_{\bm{n}}c_{\bm{n}}(t)\psi_{\bm{n}}(\bm{x}).
\end{align}

Here, the Duffing oscillator is a two-dimensional system, and the multi-index $\bm{n}$ is given by $\bm{n}=(n_1,n_2)$. The monomial dictionary function $\psi_{\bm{n}}(\bm{x})$ is given by $\psi_{\bm{n}}(\bm{x})=x_1^{n_1}x_2^{n_2}$, and the coefficient $c_{\bm{n}}(t)$ is denoted by $c(n_1,n_2,t)$. Then, Eq.~\eqref{eq:appendix_backward_equation_expansion} is rewritten as
\begin{align}
    \label{eq:appendix_backward_equation_expansion_2d}
    \frac{\partial}{\partial t}&\sum_{n_1,n_2}c(n_1,n_2,t)x_1^{n_1}x_2^{n_2}\notag\\
    &=x_2\frac{\partial}{\partial x_1}\sum_{n_1,n_2}c(n_1,n_2,t)x_1^{n_1}x_2^{n_2}\notag\\
    &\ \ +\left(-\delta x_2-\alpha x_1-\beta x_1^3\right)
    \frac{\partial}{\partial x_2}\sum_{n_1,n_2}c(n_1,n_2,t)x_1^{n_1}x_2^{n_2}.
\end{align}

We then calculate the derivatives of the monomial dictionary functions on the right-hand side of Eq.~\eqref{eq:appendix_backward_equation_expansion_2d} as
\begin{align}
    \frac{\partial}{\partial x_1}&\sum_{n_1,n_2}c(n_1,n_2,t)x_1^{n_1}x_2^{n_2} \notag\\
    &\quad=\sum_{n_1,n_2}n_1c(n_1,n_2,t)x_1^{n_1-1}x_2^{n_2} \notag\\
    &\quad=\sum_{n_1,n_2}(n_1+1)c(n_1+1,n_2,t)x_1^{n_1}x_2^{n_2},
\label{eq:appendix_backward_equation_expansion_derivatives}
\end{align}
\begin{align}
    \frac{\partial}{\partial x_2}&\sum_{n_1,n_2}c(n_1,n_2,t)x_1^{n_1}x_2^{n_2} \notag\\
    &\quad=\sum_{n_1,n_2}n_2c(n_1,n_2,t)x_1^{n_1}x_2^{n_2-1} \notag\\
    &\quad=\sum_{n_1,n_2}(n_2+1)c(n_1,n_2+1,t)x_1^{n_1}x_2^{n_2}.
\label{eq:appendix_backward_equation_expansion_derivatives_x2}
\end{align}
Substituting Eqs.~\eqref{eq:appendix_backward_equation_expansion_derivatives} and~\eqref{eq:appendix_backward_equation_expansion_derivatives_x2} into Eq.~\eqref{eq:appendix_backward_equation_expansion_2d} and equating the coefficients of each monomial $x_1^{n_1}x_2^{n_2}$, we obtain
\begin{align}
    \frac{\partial}{\partial t}c(n_1,n_2,t)
    &=(n_1+1)c(n_1+1,n_2-1,t) \notag\\
    &\quad-\delta n_2c(n_1,n_2,t) \notag\\
    &\quad-\alpha(n_2+1)c(n_1-1,n_2+1,t) \notag\\
    &\quad-\beta(n_2+1)c(n_1-3,n_2+1,t).
\label{eq:appendix_backward_equation_expansion_2d_coefficients}
\end{align}
Here, coefficients with indices outside the truncated index set are set to zero.

By solving the system of ordinary differential equations in Eq.~\eqref{eq:appendix_backward_equation_expansion_2d_coefficients}, we obtain the expansion coefficients $c(n_1,n_2,t)$. The initial condition for $c(n_1,n_2,t)$ depends on which row of the Koopman matrix is being calculated. If we calculate the $j$-th row of the Koopman matrix corresponding to the $j$-th dictionary function $\psi_j(\bm{x})=x_1^{j_1}x_2^{j_2}$, the initial condition is given by
\begin{equation}
    c(n_1,n_2,0)=
    \begin{cases}
        1, & (n_1,n_2)=(j_1,j_2),\\
        0, & \text{otherwise}.
    \end{cases}
    \label{eq:appendix_backward_equation_expansion_2d_initial_condition}
\end{equation}
This is because the time evolution of the $j$-th dictionary function is given by $\phi_j(\bm{x},t)=\sum_{\bm{n}}c_{\bm{n}}(t)\psi_{\bm{n}}(\bm{x})$, and at $t=0$, we have $\phi_j(\bm{x},0)=\psi_j(\bm{x})$.

Hence, in this work, it is necessary to solve 10 systems of ordinary differential equations corresponding to the 10 dictionary functions. The reason for choosing 10 dictionary functions is that using monomial functions up to the third order in a two-dimensional system results in a total of 10 such functions.

\section{Equivalence between the Proposed Method and the Online EDMD}
\label{app:online_edmd}
In this section, we show that the batch estimator in
Eq.~\eqref{eq:closed_form_si_edmd} can be implemented recursively by
online EDMD.\cite{zhang2019online,ohta2025integrated} For brevity, let
\begin{equation}
    \bm{p}_m=\bm{\Psi}_{\mathrm{trunc}}(\bm{x}_m),
    \qquad
    \bm{q}_m=\bm{\Psi}_{\mathrm{trunc}}(\bm{y}_m).
    \label{eq:online_edmd_lifted_snapshots}
\end{equation}
For the same regularization parameter $\lambda>0$ as in
Eq.~\eqref{eq:prior_regularized_edmd}, initialize the online algorithm by
\begin{align}
    A_0&=\lambda\widetilde K_{\mathrm{prior}}, \notag\\
    G_0&=\lambda I, \notag\\
    \widehat{K}_0&=A_0G_0^{-1}=\widetilde K_{\mathrm{prior}}.
    \label{eq:online_edmd_initial_conditions}
\end{align}
After the first $m$ snapshot pairs have been processed, define
\begin{align}
    A_m &= A_0+\sum_{i=1}^{m}\bm{q}_i\bm{p}_i^\top, \notag\\
    G_m &= G_0+\sum_{i=1}^{m}\bm{p}_i\bm{p}_i^\top, \notag\\
    \widehat{K}_m &= A_mG_m^{-1}.
    \label{eq:online_edmd_n_step_update}
\end{align}
Because $G_0=\lambda I$ with $\lambda>0$, $G_m$ is positive definite
and hence invertible for every $m$.  Upon receiving the $(m+1)$-th
snapshot pair, the sufficient statistics satisfy
\begin{align}
    A_{m+1} &= A_m+\bm{q}_{m+1}\bm{p}_{m+1}^\top, \notag\\
    G_{m+1} &= G_m+\bm{p}_{m+1}\bm{p}_{m+1}^\top, \notag\\
    \widehat{K}_{m+1} &= A_{m+1}G_{m+1}^{-1}.
    \label{eq:online_edmd_update}
\end{align}
According to Ref.~\cite{sherman1950adjustment}, the Sherman--Morrison formula gives
\begin{equation}
    G_{m+1}^{-1}
    =G_m^{-1}
    -\gamma_{m+1}G_m^{-1}\bm{p}_{m+1}\bm{p}_{m+1}^\top G_m^{-1},
    \label{eq:online_edmd_update_G_inverse}
\end{equation}
where
\begin{equation}
    \gamma_{m+1}
    =\frac{1}{1+\bm{p}_{m+1}^\top G_m^{-1}\bm{p}_{m+1}}.
    \label{eq:online_edmd_update_gamma}
\end{equation}
Substituting Eqs.~\eqref{eq:online_edmd_update_G_inverse} and
\eqref{eq:online_edmd_update_gamma} into Eq.~\eqref{eq:online_edmd_update}
yields the recursive least-squares update
\begin{align}
    \widehat{K}_{m+1}
    ={}&\widehat{K}_m
    +\gamma_{m+1}
    \left(\bm{q}_{m+1}-\widehat{K}_m\bm{p}_{m+1}\right)\bm{p}_{m+1}^\top G_m^{-1}.
    \label{eq:online_edmd_update_koopman_matrix}
\end{align}

Finally, setting $m=M$ in Eq.~\eqref{eq:online_edmd_n_step_update}
gives
\begin{align}
    A_M
    &=\lambda\widetilde K_{\mathrm{prior}}
      +\sum_{i=1}^{M}
       \bm{\Psi}_{\mathrm{trunc}}(\bm{y}_i)
       \bm{\Psi}_{\mathrm{trunc}}(\bm{x}_i)^\top, \notag\\
    G_M
    &=\lambda I
      +\sum_{i=1}^{M}
       \bm{\Psi}_{\mathrm{trunc}}(\bm{x}_i)
       \bm{\Psi}_{\mathrm{trunc}}(\bm{x}_i)^\top.
    \label{eq:online_edmd_n_step_update_final}
\end{align}
Therefore, $\widehat{K}_M=A_MG_M^{-1}$ is exactly the batch solution in
Eq.~\eqref{eq:closed_form_si_edmd}.  Thus, for any $\lambda>0$, the
proposed batch estimator is equivalent to online EDMD initialized as in
Eq.~\eqref{eq:online_edmd_initial_conditions}.  In particular, the
initialization $G_0=I$ and
$A_0=\widetilde K_{\mathrm{prior}}$ corresponds to the special case
$\lambda=1$.

\appendix
\section{Evaluation Based on One-Step State Increments}\label{app:increment_error}
In Sects.~\ref{sec:coupled_duffing_oscillators}
and~\ref{sec:coupled_van_der_pol_oscillators},
we evaluated the prediction accuracy of each method using the one-step
prediction error $\overline{e}^{\mathrm{one-step}}$.
However, because the sampling interval used in this study is relatively
small, i.e., $\Delta t=0.01$, the magnitude of the error in Eq.~\eqref{eq:one_step_prediction_error}
alone may not fully indicate how accurately the actual one-step state
evolution is reproduced. In particular, the denominator
$\|\bm{X}^{\mathrm{test},i}\|_F$ is generally of order $\mathcal{O}(1)$,
whereas the true one-step state increment is of order
$\mathcal{O}(\Delta t)$.

Therefore, as a complementary evaluation, we normalize the prediction
error by the magnitude of the true one-step state increment. We define
the matrix of the current states as
\begin{equation}
\bm{X}_{-}^{\mathrm{test},i}
=
\begin{bmatrix}
\bm{x}_{0}^{\mathrm{test},i}
&
\ldots
&
\bm{x}_{M_{\mathrm{test}}-1}^{\mathrm{test},i}
\end{bmatrix},
\end{equation}
and define the relative one-step increment error for test trajectory $i$ as
\begin{align}
e_{\mathrm{inc}}^{(i)}
&=
\frac{
\left\|
\left(\widehat{\bm{X}}^{\mathrm{test},i}
      -\bm{X}_{-}^{\mathrm{test},i}\right)
-
\left(\bm{X}^{\mathrm{test},i}
      -\bm{X}_{-}^{\mathrm{test},i}\right)
\right\|_F
}{
\left\|
\bm{X}^{\mathrm{test},i}
-\bm{X}_{-}^{\mathrm{test},i}
\right\|_F
}
\notag\\
&=
\frac{
\left\|
\widehat{\bm{X}}^{\mathrm{test},i}
-\bm{X}^{\mathrm{test},i}
\right\|_F
}{
\left\|
\bm{X}^{\mathrm{test},i}
-\bm{X}_{-}^{\mathrm{test},i}
\right\|_F
}.
\label{eq:one_step_increment_error}
\end{align}
The numerator represents the discrepancy between the predicted and true
one-step state increments, whereas the denominator represents the magnitude
of the true one-step state increments. Thus, smaller values indicate more
accurate prediction of the one-step state evolution, and
$e_{\mathrm{inc}}^{(i)}=0$ corresponds to perfect prediction.

For each random seed, we average this error over all test trajectories:
\begin{equation}
\overline{e}_{\mathrm{inc}}
=
\frac{1}{N_{\mathrm{test}}}
\sum_{i=1}^{N_{\mathrm{test}}}
e_{\mathrm{inc}}^{(i)}.
\end{equation}

\begin{figure}
    \centering
    \includegraphics[width=\linewidth]
    {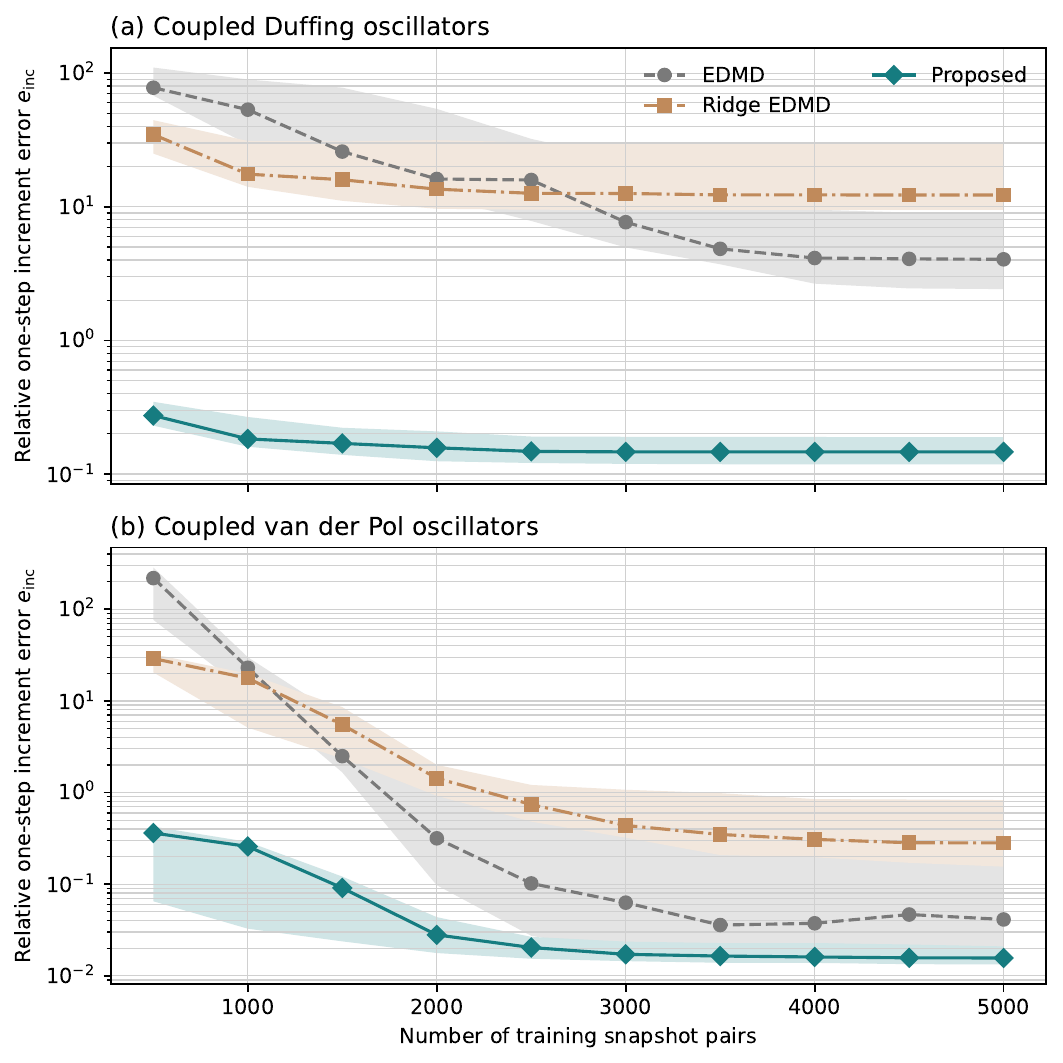}
    \caption{(Color online) The median and interquartile range of the
    relative one-step increment error $\overline{e}_{\mathrm{inc}}$
    across $20$ independent random seeds for
    (a) the coupled Duffing system and
    (b) the coupled van der Pol system.
    The horizontal axis represents the number of training snapshot pairs.
    The lines and shaded regions represent the medians and interquartile
    ranges, respectively.}
    \label{fig:one_step_increment}
\end{figure}

Figure~\ref{fig:one_step_increment} shows the relative one-step increment
errors for the two coupled systems. For both systems, the proposed method
yields lower increment errors than standard EDMD and
$\ell_2$-regularized EDMD. In addition, the error of the proposed method
remains substantially below unity, indicating that its prediction error
is smaller than the magnitude of the actual one-step state increment.
These results are consistent with the state-normalized one-step prediction
errors in Figs.~\ref{fig:duffing_one_step}
and~\ref{fig:vdp_one_step}, and confirm that the advantage of the proposed
method is not merely a consequence of the small sampling interval.
\bibliographystyle{apsrev4-2}
\bibliography{main}
\end{document}